\pdfoutput=1

\documentclass[11pt]{article}

\usepackage[final]{acl}

\usepackage{xcolor}

\definecolor{theindigo}{RGB}{113, 10, 186}

\usepackage{times}
\usepackage{latexsym}
\usepackage{amsfonts,amssymb}
\usepackage[fleqn]{amsmath}
\usepackage{stmaryrd}  %
\usepackage{mathtools}
    \mathtoolsset{showonlyrefs, showmanualtags, mathic}
\usepackage{mleftright}
\usepackage{subcaption}
\usepackage{xurl}

\usepackage[T1]{fontenc}

\usepackage[utf8]{inputenc}

\usepackage{microtype}

\usepackage{inconsolata}

\usepackage{graphicx}

\usepackage{booktabs}
\usepackage{fnpct}  %
\usepackage{csquotes}
\usepackage{cleveref}
\usepackage{siunitx}
	\DeclareSIUnit{\quantity}{\relax}
	\DeclareSIUnit{\words}{words}
	\DeclareSIUnit{\sentences}{sentences}
\usepackage{tabularray}
    \UseTblrLibrary{booktabs}

\usepackage{tikz}
    \usetikzlibrary{arrows}
	\usetikzlibrary{matrix, chains, graphs}
	\usetikzlibrary{positioning, calc}
    \usetikzlibrary{bayesnet}

\title{A Bayesian account of pronoun and neopronoun acquisition}

\author{Cassandra L. Jacobs \\
  Department of Linguistics \\
  State University of New York at Buffalo\\
  Buffalo, NY, USA \\
  \texttt{cxjacobs@buffalo.edu} \\\And
  Morgan Grobol \\
  MoDyCo \\
  Université Paris Nanterre \\
  Nanterre, France \\
  \texttt{lgrobol@parisnanterre.fr} \\}

\begin{document}
\maketitle
\begin{abstract}
A major challenge to equity among members of queer communities is the use of one's chosen forms of reference, such as personal names or pronouns.
Speakers often dismiss their misuses of pronouns as \enquote{unintentional}, and claim that their errors reflect many decades of fossilized mainstream language use, as well as attitudes or expectations about the relationship between one's appearance and acceptable forms of reference.
We argue for explicitly modeling individual differences in pronoun selection and present a probabilistic graphical modeling approach based on the nested Chinese Restaurant Franchise Process (nCRFP) \cite{ahmed2013nested} to account for flexible pronominal reference such as chosen names and neopronouns %
while moving beyond form-to-meaning mappings
and without lexical co-occurrence statistics to learn referring expressions, as in contemporary language models.
We show that such a model can account for variability in how quickly pronouns or names are integrated into symbolic knowledge and can empower computational systems to be both flexible and respectful of queer people with diverse gender expression.
\end{abstract}

\section{Introduction}

In contrast to words that are used to label referents as determined by convention (e.g., \enquote{cat} refers to \textsc{cat}-like entities; \citealp{brennan1996conceptual}), people have the autonomy to change their names and update their pronouns to reflect their identity \cite{Zimman+2019+147+175}.
In many Western cultures, however, personal names and pronouns are usually assigned \emph{to} someone \emph{by} others (e.g., one's parents or the norms of the ambient culture; \citealp{lind2023gender}), and are highly conventionalized.
For example, English canonically has only two animate third-person singular pronouns (i.e., he/him/his and she/her/hers).
These pronominal forms as well as personal names are strong cues to gender identity.
Within linguistics, this regularity has led to the general practice of treating referring expression generation as a form-to-meaning mapping problem \cite{enfield2007person}.
That said, the forms of reference used for someone are neither fixed, nor intrinsic properties of an individual.
This paper presents a probabilistic modeling framework that respects a person's right to self-determination (of how to be referred to) without positing form-to-meaning or form-to-feature mappings.
Our proposal accounts for the ongoing sociolinguistic change among young Westerners to ask and reinforce their understanding of their peers' self-identities.

The need for modeling pronoun and name use in natural language processing (NLP) is especially important given the increasing prominance of accommodating individuals' identities in the public sphere.
Despite major advances in natural language generation, it has proven difficult to incorporate this into modern systems, especially in present-day neural network models.
For example, even the most basic rule-based tokenization systems still do not flexibly handle nonbinary forms of address such as \enquote{Mx.}
Furthermore, large language models (LLMs) and commercial generative AI systems perpetuate bias against women and gender minorities by encoding harmful stereotypes in their training data (e.g., negative sentiment; \citealp{dev-etal-2021-harms,ungless-etal-2023-stereotypes}) for in marginalized individuals' names, common professions, personal items, and pronouns.
This is even more true for queer people outside the gender binary, as datasets regularly exclude nonbinary identities from their construction \cite{hall2023visogender,sakaguchi2021WinoGrandeAdversarialWinograd}.
Language that does not conform to gender stereotypes is also mishandled by NLP systems \cite{10.1145/3600211.3604672,havens-etal-2022-uncertainty}.

Here, we propose that systems that symbolically encode valid referring expressions for individuals are less prone to these problems.
With present limitations in mind, we outline below the basic capabilities of an ideal system for learning the forms and representations of an individual's referring expressions such as names and pronouns must include:
\begin{enumerate}
    \item Allow the introduction of new forms into the vocabulary (e.g., novel names or neo-pronouns)
    \item Permit individuals to use a mixture of forms of reference for themselves (e.g., alternating between he/she/they or using different gendered forms in different languages; \citealp{moore2024queer})
    \item Quickly adapt in the face of revision (e.g., updates to a person's name or pronouns), potentially given a single exemplar
    \item Allow adaptation to vary by individuals
\end{enumerate}

We further argue that such a system should produce more flexible adaptation for individuals who are more accustomed to such adaptation.

\section{A Dirichlet process model of name and pronoun learning}

Due to its symbolic nature, our proposed system can learn appropriate forms of address and reference through experience without encoding discriminatory knowledge such as an individual's appearance into their representations.
This empowers queer people and supports their autonomy \cite{lind2023gender,10.1145/3593013.3594078,Zimman+2019+147+175}.
We treat the learning process as the assignment of probabilities of referential forms -- pronominal or otherwise -- directly to individuals rather than through the medium of individual characteristics \cite{lauscher-etal-2022-welcome}.

\begin{figure}[th]
    \centering
    \begin{tikzpicture}[>=stealth]
        \node[obs] (prot) {\(\mathrm{pro}^t\)}; 
        \node[obs, left=of prot] (u) {\(u\)};
        \node[latent, left=of u] (T) {\(T\)};
        \node[latent, above=of prot] (P) {\(P\)};
        \node[obs, above=of P] (proo) {\(\mathrm{pro}^o\)};
        \node[latent, right=of prot] (Pt) {\(P^t\)};

        \plate {interaction} {(u)(prot)} {\(I\)};
        \plate {discourse} {(T)(interaction)(Pt)} {\(D\)};

        \plate {} {(proo)} {\(N\)};
     
        \graph [use existing nodes] {
            T -> u,
            u -> prot,
            prot ->[loop below] prot,
            proo -> P,
            P -> prot,
            Pt -> prot,
        };
    \end{tikzpicture}
    \caption{Single speaker model}\label{fig:speaker}
\end{figure}
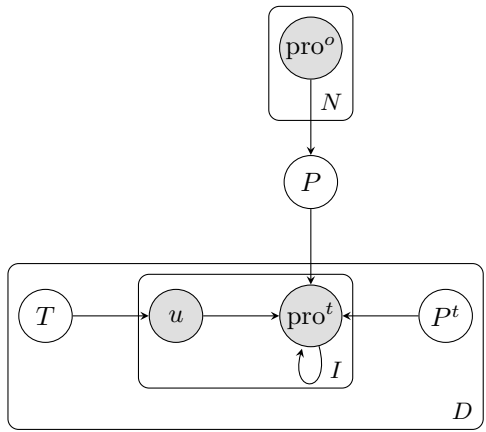

Latent Dirichlet Allocation  (LDA; \citealp{NIPS2001_296472c9}) is an algorithm that allows the probabilistic assignment of discrete labels (e.g., topics) to collections of events (e.g., documents) on the basis of the contents of the document (e.g., words).
As suggested by the name, the topics learned by LDA are latent variables that are unobservable.
In this modeling framework, documents are observable objects that are assumed to be generated by sampling words from mixtures of topics.
Critically, a trained topic model can be used to estimate what proportion of topics was used to generate that document.
These models are in principle infinite, and can have novel topics as well as additional vocabulary items added as a dimension in the vocabulary by trivial extension.

Building on this approach, the nested Chinese Restaurant Franchise Process (nCRFP; \citealp{ahmed2013nested}) allows for models to even learn that different types of documents or users exist.
For example, book chapters and magazine articles may have different lexical distributions, and authors within each of those genres may have different lexical preferences.
Graphical models have been used to capture variation in language use across different geographical regions \cite{eisenstein2010LatentVariableModel} -- analogous to the speaker communities of interest here.
Simplified versions of Dirichlet processes (e.g., Beta-binomial priors) have also been applied to learning, as in learning and adaptation to syntactic structures in the context of a conversation \cite{kleinschmidt2012belief}.

The present paper expands the metaphor of the nCRFP \cite{ahmed2013nested} to model an individual's learning of referring expressions -- and specifically the pronouns -- for others.
We choose to treat pronouns or similar gender markers as observable objects that have probabilistic assignment to topics (communities of individuals), making pronouns most analogous to words in a document.
Furthermore, we can characterize individuals or referents as \enquote{documents} that comprise a unique probability distribution over pronouns and names.
Extending the metaphor to the hierarchical domain, different communities of learners (topics) may have priors of different strengths and/or more uniform expectations over pronoun use for unfamiliar individuals. 
Within topics, it is also clear that different groups of learners belong to different communities that reinforce the statistics of use of referring expressions within their communities.

\section{Probabilistic graphical model of individual speaker preference}

In \Cref{fig:speaker}, we present the parametrization of the single-speaker model, which details how a speaker selects pronouns referring to a specific individual \(t\) across utterances as a function of their linguistic experience. This model involves the following variables (indices are omitted in the figure for brevity):

\begin{description}
    \item[\(\mathrm{pro}^t_{d, i}\)] Produced pronoun referring to \(t\) in the  interaction \(i\) of discourse \(d\). Can be absent, in case where the preferred pronouns are no pronouns. The self-loop allows for both pronoun stability and intentional alternation. That is, speakers can either select a chosen pronoun for a particular interaction, which they adhere to, or vary pronoun uses if the referent has indicated such a preference.
    \item[\(u_{d,i}\)] Utterance including a pronominal reference to \(t\).%
    \item[\(P\)] The speaker's general prior on pronoun production.
    \item[\(P^t_d\)] The speaker's prior on \(t\)'s pronouns at the time of interaction \(d\). The support of \(P^t_d\) is not necessarily pointwise, and its support and distribution are subject to adjustments between different interactions, for instance in case of offline feedback about a pronoun use.
    \item[\(T_d\)] Topic for interaction \(d\).
    \item[\(\mathrm{pro}^o_n\)] Pronoun usages witnessed by the speaker at all times and for any referent.
\end{description}

These variables are plated across the set \(D\) of all discourses (spoken or written) where the speaker has referred to \(t\), the set \(I\) of all interactions in said discourse, and the set \(N\) of all interactions witnessed at all by the Speaker.

A Bayesian approach captures the intuition that some individuals may have more rigid \enquote{priors} over pronouns for specific speakers, and therefore choose to override the referent's choice of pronouns. While this relative stubbornness is expected among individuals who adhere to gender binaries, it could also arise in individuals who are willing to expand their pronominal inventory but struggle to do so without significant exposure to more diverse pronoun usages. 

Note that our models do not assume any reliance on external characteristics. While we generally disagree with the practice, a speaker's prior belief over pronoun distribution could be jointly determined by both linguistic experience as well as the co-occurrence of such characteristics in order to account for intentional or unintentional misgendering.

\section{Probabilistic graphical model of community norms}

\begin{figure*}[th]
    \centering
    \begin{tikzpicture}[>=stealth]
        \node[obs] (prot) {\(\mathrm{pro}^{s,t}\)}; 
        \node[obs, left=of prot] (u) {\(u\)};
        \node[latent, left=of u] (T) {\(T\)};
        \node[latent, above=4em of prot] (P) {\(P^c\)};
        \node[latent, right=5em of prot] (Pt) {\(P^{c,t}\)};

        \plate[inner sep=0.75em]{interaction} {(u)(prot)} {\(I\)};
        \plate[inner sep=1em] {discourse} {(T)(interaction)} {};
        \plate {referent} {(prot)(Pt)(interaction.south)(interaction.north east)} {};
        \plate {speaker} {(interaction)(discourse)(P)(referent)} {};
        \node[caption, node distance=0, inner sep=0pt, below left=1em and 1em of speaker.north east] {\(c\in C\)};
        \node[caption, node distance=0, inner sep=0pt, above right=0.4em and 0.5em of discourse.south west] {\(D\)};
        \node[caption, node distance=0, inner sep=0pt, above left=0.4em and 0.5em of referent.south east] {\(t\in C\)};

        \graph [use existing nodes] {
            T -> u,
            u -> prot,
            prot ->[loop below] prot,
    
            P ->[
                bend left,
                edge node={
                    node[near start, sloped, auto]{
                        \scriptsize\(c=s\)
                    }
                }
            ] prot,
            prot ->[bend left] P,
            Pt ->[
                bend left,
                edge node={
                    node[near start, sloped, auto]{
                        \scriptsize\(c=s\)
                    }
                }
            ] prot,
            prot ->[bend left] Pt,
        };
    \end{tikzpicture}
    \caption{Community model with many speakers.}%
    \label{fig:community} %
\end{figure*}
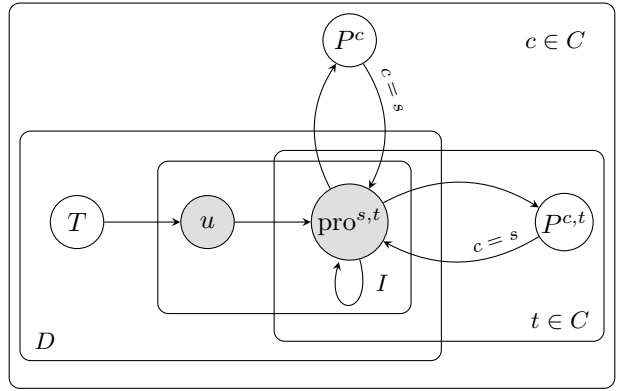

Speakers do not obtain their linguistic knowledge from pure distributional statistics. 
Rather, their preferences are contextualized by interactions with others in their language communities and through interactions with individuals that may reinforce those community beliefs.
In cases where a speaker belongs to a community with practices that either accept and embrace --- or deny --- the practice of naming oneself \cite{lind2023gender}, speaker priors are expected to be sampled from the community prior over pronouns as well.

For example, queer and cis-binary communities display clear differences in linguistic preferences and consensus about whether one's pronouns neatly correspond to one's current presentation suggests \cite{rose2023VariationAcceptabilityNeologistic}.
This gives rise to the prediction that some speakers will not readily adapt to signals that (in a given conversation) the relevant pronouns to use belong to some set and not others \cite{arnold2024gender}, particularly if their linguistic knowledge strictly excludes gender neutral or neopronouns.
On the other hand, queer folks who have many friends whose pronouns fall outside the gender binary can be expected to have more flexible and more uniform beliefs about potential pronouns.

At the scale of a whole community, where pronoun usage witnessed by someone are those produced by other members of a the community, our model becomes that of \Cref{fig:community}: for all triplets \(c, s, t\) of individuals in a community \(C\), \(\mathrm{pro}^{s,t}\) is a pronoun used by a \(s\) to refer to \(t\) and \(P^{c,t}\) and \(P^c\) are the priors of \(c\) about possible pronoun usages, respectively for \(t\) specifically, and for anyone. Note that the self-referring case \(s=t\) is not excluded, and is in fact an important part in building priors for the rest of the community.

\section{Related work}
A challenge for modeling pronoun use in practical systems arises when we presuppose that learning words boils down to the problem of mapping form onto meaning.
For instance, early connectionist approaches to semantic representation, have treated the "meaning" of a word as a sparse $d$-dimensional vector consisting of several manually-selected semantic features \cite{cree2006distinctive,10.7551/mitpress/5236.001.0001}.
Here, we propose that meaning be defined symbolically at the level of a referent rather than distributed across semantic features.

In word vectors trained on corpora, a \enquote{gender subspace} commonly emerges \cite{NIPS2016_a486cd07} that encodes social biases about canonical genders (e.g., stereotypes about the gender of nurses versus doctors).
Pronouns and other high-frequency gendered nouns (e.g., man, woman) typically serve as critical anchors in the debiasing process, and serve as an excellent probe into the origins of biases in modern statistical NLP systems.
Others have successfully demonstrated that non-binary pronoun LLM representations can be debiased, suggesting that the form-to-meaning mapping can be partially undone for novel referential forms \cite{vanboven2024TransformingDutchDebiasing}.

Being able to appropriately select the correct pronoun for a referent, as in text generation applications, is critical for ensuring equity and access to modern-day NLP tools.
A number of studies have attempted to study gender bias in pronoun production.
However, few of these studies have been able to probe the representations of pronouns,  neopronouns, and name use that differs from the mainstream \cite{sakaguchi2021WinoGrandeAdversarialWinograd}.
The model we present here is capable of generating a wide variety of potential sentences to test the role of experience during fine-tuning of language models and thus improve gender inclusivity.

The present work is strongly informed by the integrative account presented in \citet{ackerman2019syntactic}, who stated that cognitive, biological, and social factors combine to influence coreference resolution for non-binary people.
They highlight that normatively unexpected mappings can nevertheless be made felicitous with sufficient supporting context.

\section{Future Work}

Our models allow for a straightforward integration of both witnessed pronoun uses and external priors in the process of pronoun selection in production. This provides a reasonably simple way to model pronoun acquisition during a long history of interactions in communities. However, for the sake of simplicity, certain interaction dynamics are not taken into account, and we leave to future work the search for improved models that balance the insights added by these refinements and the extra complexity that they would induce.

Our community model does not explicitly include non-linguistic social dynamics. Most importantly, language uses witnessed by a comprehender might have different weights depending on the speaker. For instance, the credit given to pronoun uses by speaker \(s\) for referent \(t\) could vary depending on how close to \(t\) \(s\) is assumed to be, and the \(s=t\) case could be given a separate treatment.
Furthermore, our models are only concerned with pronouns, which have the lightest semantic content of all referring expressions. However, it is likely that in practice, pronoun usage is also informed by the use of other referring expressions, such as names, formal titles, terms of address, etc. In our current model, these evidences are folded into the priors \(P\), but more precise examination of their internal structure would provide a much richer model.

\section{Conclusion}

The model that we outline here shows that it is possible to represent individuals as possessing distributions of pronominal referring expressions, consistent with their own self-determined gender.
The probabilistic graphical modeling accounts are flexible enough to allow learners to accommodate others based on their experience with linguistic variability in pronoun use.
Additionally, the work provides a mechanism for the easy extension of one's linguistic vocabulary to incorporate novel pronouns, including but not limited to neopronouns, emojipronouns, and so on.
We view this work as a critical bridge between cognitive scientific work on pronoun processing \cite{ackerman2019syntactic,arnold2024gender,rose2023VariationAcceptabilityNeologistic} and computational modeling of linguistic variability \cite{eisenstein2010LatentVariableModel,kleinschmidt2012belief} while also providing a way to advance equity in pronoun generation and comprehension \cite{10.1145/3593013.3594078,piergentili-etal-2024-enhancing,lauscher2023WhatEmHow}.

\bibliography{custom}

\end{document}